\documentclass{article}

\usepackage[preprint]{neurips_2025}
\usepackage[utf8]{inputenc}
\usepackage[T1]{fontenc}
\usepackage{amsmath,amssymb}
\usepackage{graphicx}
\usepackage{booktabs}
\usepackage{tabularx}
\usepackage{array}
\usepackage{enumitem}
\usepackage{float}
\usepackage{microtype}
\usepackage{xcolor}
\usepackage{hyperref}
\usepackage{url}

\hypersetup{
  colorlinks=true,
  linkcolor=blue!55!black,
  citecolor=blue!55!black,
  urlcolor=blue!55!black
}

\setlist{leftmargin=*,itemsep=2pt,topsep=3pt}

\newcommand{\Pexpected}{P_{\mathrm{expected}}}
\newcommand{\Pobserved}{P_{\mathrm{observed}}}
\newcommand{\GammaH}{\Gamma_{\mathrm H}}
\newcommand{\Nwork}{N_{\mathrm{work}}}
\newcommand{\Hdep}{H_{\mathrm{dep}}}

\title{Benchmarking the Residual:\\
What Long-Horizon Evaluations Add Beyond\\
Matched Short-Task Performance}
\author{%
  Chao Peng\thanks{These authors contributed equally.}%
  \hspace{0.55em}\thanks{Corresponding authors: \texttt{chao.peng@acm.org} and \texttt{handedong@tencent.com}.}%
  \quad Zhiheng Lyu\footnotemark[1]%
  \quad Peijie Dong\footnotemark[1]%
  \quad Hande Dong\footnotemark[2]%
  \quad Qiang Lin\\[-0.1em]
  {\normalfont Tencent}\\[-0.2em]
  {\normalfont\footnotesize\ttfamily
    chao.peng@acm.org, zl149@illinois.edu, peytondong@tencent.com}\\[-0.2em]
  {\normalfont\footnotesize\ttfamily
    handedong@tencent.com, cheaterlin@tencent.com}
}

\begin{document}
\raggedbottom
\maketitle
\setcounter{footnote}{0}

\begin{abstract}
Long-horizon benchmarks often show that agents fail more as tasks become longer. This observation is useful for deployment, but it does not by itself explain why failure occurs. More stages create more opportunities for ordinary errors to compound; longer tasks may also contain harder individual decisions or become harder as conversation history, tool outputs, and environment changes accumulate. We use \emph{trajectory-induced degradation} to mean this last possibility: earlier execution makes later work harder. When the harmful accumulation is specifically the text visible to the model, it is often called \emph{context rot}. In this position paper, we argue that to claim a ``long-horizon failure'', benchmarks must compare actual full-task success against a baseline prediction built from short, individual stages. We call the log-ratio between this prediction and actual success the \emph{horizon residual}. The comparison must use the same agent configuration and specify in advance how stages, checkpoints, information, and budgets will be chosen. The residual shows that the full rollout differs from the chosen baseline; targeted experiments are still needed to explain why.
\end{abstract}

\begin{center}
\begingroup
\setlength{\fboxsep}{5pt}
\setlength{\fboxrule}{0.7pt}
\fcolorbox{blue!55!black}{blue!4}{\begin{minipage}{0.93\linewidth}
\small
\textcolor{blue!55!black}{\textbf{The key idea.}}
\textbf{Benchmark~design:} Longer can mean more work, not harder work. Benchmarks should distinguish task size from how hard each stage is and how strongly stages depend on one another. \textbf{Result~interpretation:} More stages create more chances to fail. Compare actual full-task success with a prediction built from matched short stages; the \emph{horizon residual} summarizes the mismatch, not its cause.
\end{minipage}}
\endgroup
\end{center}

\section{Introduction}

Language-model agents are increasingly evaluated on tasks that extend beyond a single response or isolated code edit. In software engineering, recent benchmarks range from repository-level issue resolution~\citep{jimenez2024swebench} to release-sized evolution~\citep{le2025sweevo}, chained package upgrades~\citep{lam2026swechain}, and continuous milestone execution on a persistent codebase~\citep{deng2026swemilestone}. Related interactive benchmarks require agents to navigate websites, operate desktop applications, coordinate APIs, or conduct iterative research engineering~\citep{zhou2023webarena, xie2024osworld,trivedi2024appworld,wijk2024rebench}. Across these settings, success generally becomes less reliable as the amount of work, the depth of inter-stage dependence, and the accumulated execution history increase. This trend matters for deployment because many valuable tasks require sustained planning, tool use, state tracking, verification, and recovery over extended trajectories.

The phrase \emph{long horizon}, however, can refer to several different quantities:
\begin{itemize}
  \item the minimum number of effective actions required;
  \item the number of separately verifiable milestones;
  \item the depth of the dependency chain;
  \item the duration of the agent's rollout; or
  \item the amount of prior history visible to the model.
\end{itemize}
These quantities often move together in naturally collected tasks, but they need not. A task may contain many independent operations with shallow dependencies, or only a few stages that strongly depend on earlier state. A short task may also produce a long trace because the agent retries or explores irrelevant actions. Treating these cases as one notion of length obscures what a benchmark actually measures~\citep{wang2026mirage,kwa2025longtasks}.

\textbf{A lower end-to-end success rate on a longer task is therefore not yet a diagnosis of a long-horizon failure mechanism.} Even if each local stage is unchanged, requiring more stages to succeed can reduce task success through ordinary error compounding. Longer tasks may also contain intrinsically harder local decisions, expose the agent to noisier context, or allow earlier actions to alter the state on which later decisions depend. Existing results illustrate these possibilities: independent and continuous milestone protocols can produce sharply different outcomes~\citep{deng2026swemilestone}; agents can degrade when they inherit their own prior code changes~\citep{jin2026chainswe}; and controlled history manipulations show sensitivity to erroneous prior outputs at a matched evaluation turn ~\citep{sinha2026illusion}. Because these factors often vary together, a raw performance curve can establish a deployment limitation without identifying its source.

This distinction separates two legitimate purposes of evaluation. A deployment benchmark asks whether the complete system succeeds under realistic budgets and interaction rules. A diagnostic benchmark asks which controlled contrast explains behavior that a simpler model of local competence does not. The first purpose prioritizes ecological validity; the second requires replayable states, explicit interventions, and assumptions that can be audited. Neither replaces the other. In practice, however, long-horizon results are often given diagnostic readings that the deployment observation alone cannot license.

We therefore propose a simple diagnostic comparison. First, run the same agent on short, verifiable stages from declared checkpoints. Next, combine those stage results into a prediction for the full task. Finally, compare the prediction with success in a natural end-to-end rollout. We use \emph{counterfactual} for this pre-specified alternative evaluation protocol, not for an individual-level causal claim.\footnote{The resulting quantities compare evaluation protocols; they are not individual-level potential-outcomes counterfactuals.} The horizon residual summarizes how far the natural rollout departs from the checkpoint-based prediction. It is not a causal answer, but it can direct follow-up experiments on history, state, planning, verification, or recovery.

This position paper makes three contributions:
\begin{enumerate}
  \item We \textbf{advocate} treating raw performance decline as a descriptive deployment result and the counterfactual residual as a separate, protocol-dependent diagnostic.
  \item We \textbf{propose} a deliberately simple and auditable product baseline, together with conditional-model and checkpoint-simulation extensions when its assumptions fail.
  \item We \textbf{articulate} reporting requirements, scope conditions, and an intervention agenda for turning residual patterns into testable causal hypotheses.
\end{enumerate}

\section{Diagnosing Performance Decline Beyond ``Longer Is Harder''}
\label{sec:motivation}

Consider a coding agent tasked with making four independent repairs. Suppose it succeeds on each repair with probability 80\% when evaluated separately. If these outcomes compose independently, then even when no additional long-horizon failure occurs, the probability of completing all four repairs is only $0.8^4\approx41\%$. Thus, a low end-to-end success rate is not by itself evidence of a distinctive long-horizon limitation; it may follow from ordinary error compounding.

Now suppose the same agent completes the four-repair task only 10\% of the time. The gap between 41\% and 10\% is more informative. Earlier edits may contaminate the repository state, long tool transcripts may distract the model, or the agent may lose track of the original goal, omit intermediate verification, or fail to recover from an early mistake. These are plausible explanations, but the raw 10\% success rate cannot resolve which, if any, is responsible. Interpretation therefore requires a declared model of what matched short-stage performance predicts.

Table~\ref{tab:three-reasons} summarizes three effects that are easy to conflate:

\begin{table}[H]
\caption{Distinct sources of performance degradation as task length increases, together with the control required to distinguish each source.}
\label{tab:three-reasons}
\centering
\small
\setlength{\tabcolsep}{4pt}
\begin{tabularx}{\linewidth}{>{\raggedright\arraybackslash}p{0.19\linewidth}XX}
\toprule
\textbf{Effect} & \textbf{What changes?} & \textbf{Useful control} \\
\midrule
Error compounding &
The local stages are no harder, but more of them must succeed. &
Measure each matched stage and compose the estimated success probabilities. \\
Harder local work &
Longer tasks contain more ambiguous, specialized, or poorly verified stages. &
Match the distribution of local stage difficulty. \\
History dependence &
Later steps become harder as context, state, or earlier errors accumulate. &
Compare retained history with reset, compressed, or repaired history. \\
\bottomrule
\end{tabularx}
\end{table}

Hereafter, \emph{trajectory-induced degradation} means that accumulated execution (the transcript, tool outputs, environment state, or earlier errors) makes later work harder. \emph{Context rot} is the narrower case in which the growing visible text is the source of the decline~\citep{hong2025contextrot}; positional degradation in long contexts is a related observation~\citep{liu2024lost}. These labels describe where performance changes, not the underlying mechanism. They may cover distraction from a noisy transcript, cross-module planning, state contamination, or failed recovery.

Recent evidence shows why these categories should not be collapsed. In a controlled running-state task, \citet{sinha2026illusion} hold the evaluation turn and history format fixed while varying the error rate of the preceding outputs. Cleaner histories improve subsequent accuracy, showing that the content of the execution history matters beyond its length. A related content-versus-length control appears in repeated social-dilemma games: holding prompt length fixed while replacing the visible history with synthetic cooperative records substantially restores cooperative behavior, isolating history content rather than length as the trigger~\citep{liu2026memorycurse}. ChainSWE provides a complementary repository-level comparison: across 100 chains containing 304 issues from 54 projects, average per-bug accuracy falls from 58.9\% when each bug begins from an oracle-correct repository state to 36.5\% when the agent inherits its own prior modifications~\citep{jin2026chainswe}. Moreover, 318 of 663 downstream sequential failures in its baseline setting occur on bugs that the same systems solve from the oracle state. These results do not establish a universal horizon law, but they illustrate the diagnostic value of comparing natural execution against a matched state-repaired counterfactual.

\section{A System-Matched Counterfactual for Long-Horizon Evaluation}
\label{sec:counterfactual}

\subsection{Decomposition and Checkpoints}

We propose beginning with a pre-specified decomposition of the long task into verifiable stages. In a software engineering task, these stages might be ``find the bug'', ``change the data model'', ``update the API'', and ``pass the integration tests''. Each stage should be semantic: it should be defined by an acceptance condition rather than by an arbitrary number of turns or tool calls.

\begin{figure}[t]
  \centering
  \includegraphics[width=\linewidth]{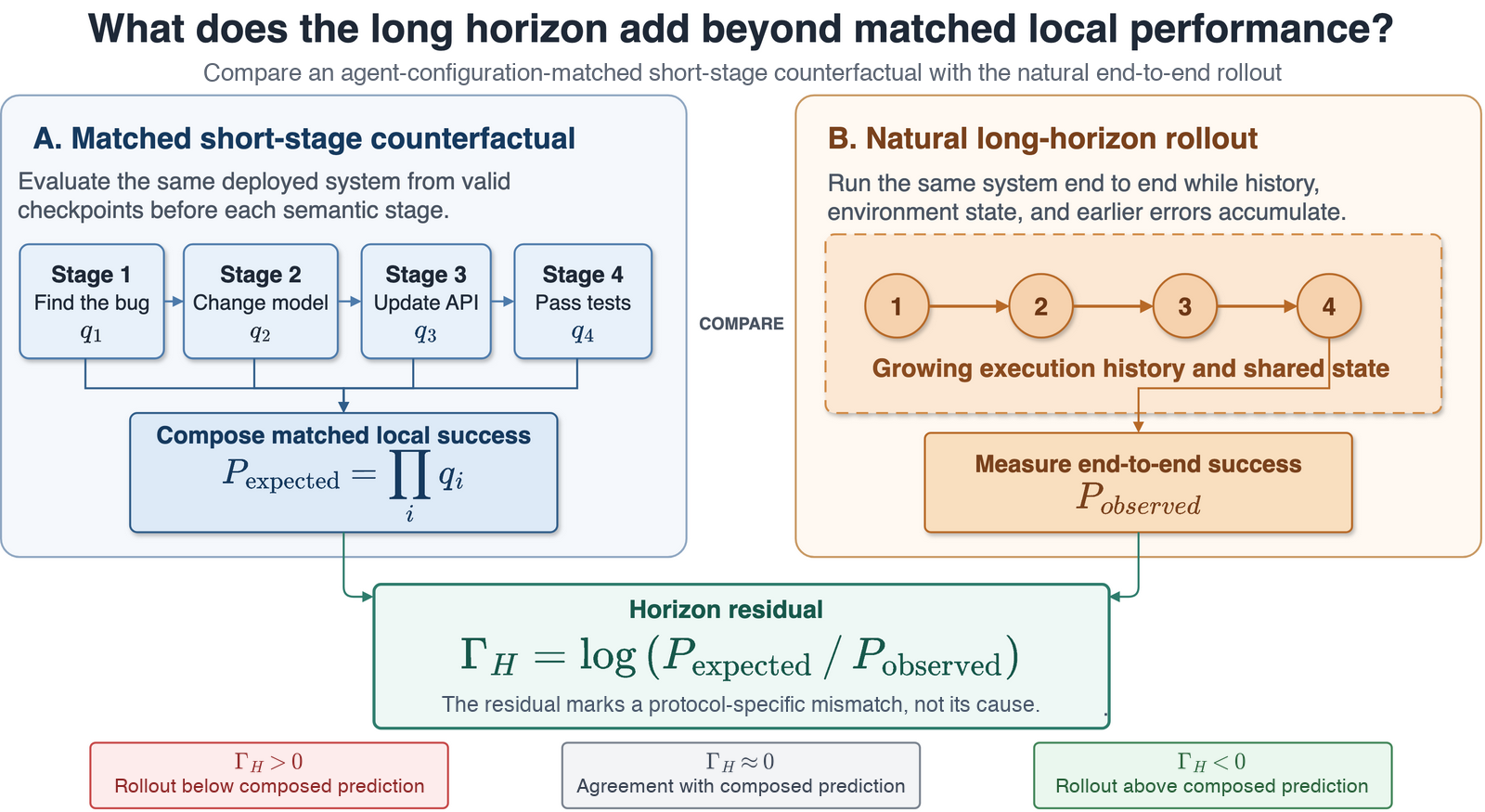}
  \caption{Agent-configuration-matched comparison between checkpoint-based local evaluation and natural end-to-end execution. Declared checkpoint interventions estimate the conditional stage probabilities used to construct $\Pexpected$, whereas the natural rollout yields $\Pobserved$ under accumulated history and shared state. The horizon residual $\GammaH$ records their protocol-specific log contrast; mechanism claims require additional interventions.}
  \label{fig:horizon-residual}
\end{figure}

The next step is to run the same deployed agent from declared checkpoints (Figure~\ref{fig:horizon-residual}). Here, the deployed agent denotes the complete system, including the model, prompt, tools, memory, and execution framework. Let $S_i$ denote success at stage $i$, and let $I$ denote a pre-specified checkpoint protocol: the environment state, visible history, revealed information, and local budget supplied at that stage. Along a fixed path on which every stage is required, define $q_i=\Pr(S_i=1\mid S_{<i}=1,I)$. Here, $q_i$ is the agent's chance of passing stage $i$ when the required earlier stages are already complete and the benchmark starts the agent from its declared checkpoint.

\subsection{The Product Baseline}

The chain rule then gives the auditable baseline

\begin{equation}
  \Pexpected = \prod_i q_i.
  \label{eq:expected}
\end{equation}

In simple terms, if an agent has a 90\% chance of passing Stage~1 and an 80\% chance of passing Stage~2 from clean, compatible states, we predict a 72\% overall success rate ($0.90\times0.80$). If the actual agent succeeds only 30\% of the time, the intuitive shortfall is 42 percentage points; the horizon residual defined below reports this same comparison as a log-ratio.

The homogeneous special case clarifies why this baseline matters. If every stage succeeds with a constant probability $p$, failures are independent, and no recovery is possible, then an $n$-stage task succeeds with probability $p^n$. Equivalently, the first horizon at which success falls below a threshold $s$ is approximately $H_s=\lceil \log(s)/\log(p) \rceil$~\citep{sinha2026illusion}. This analysis shows that a substantial raw decline can arise without any additional long-horizon mechanism. Its assumptions are deliberately strong: stage accuracy may change with position, errors may alter future state, and natural rollouts may recover. We therefore use $p^n$ as an auditable null model rather than a description of how agents necessarily behave.

Three baseline forms cover different task structures. The \emph{product baseline} multiplies conditional stage-success rates when the checkpoints form one compatible path. An \emph{oracle-scaffolded baseline} multiplies success rates from separately chosen, clean canonical states; because those states may not arise in one real rollout, it requires extra compatibility and independence assumptions. A \emph{checkpoint simulator} is needed when solutions branch, failures can be retried or recovered from, or one stage changes the state and odds of later stages.

For a heterogeneous task family, predictions should be composed within each task and then aggregated using the same task weights as $\Pobserved$; multiplying stage averages can give a different quantity. In every case, the benchmark should declare the baseline model, estimate it from the corresponding checkpoint behavior, and compute its prediction before evaluating the natural end-to-end rollout.

Checkpoint evaluation also raises a cross-world composition problem. An oracle reset, repair of an agent-produced state, fresh context, and revelation of a local goal are separate interventions and therefore define different estimands.

Moreover, an alternative valid solution at stage $i$ need not be composable with a canonical downstream checkpoint constructed from a reference solution. A benchmark should declare whether it evaluates a canonical path, constructs solution-conditioned downstream states, or simulates transitions across compatible states; otherwise the $q_i$ values may describe stages that cannot be combined in one coherent rollout.

Existing results suggest that checkpoint and continuous protocols can differ sharply. SWE-Milestone evaluates 98 human-verified milestones across seven repositories and five programming languages. The reported scores are approximately 80\% or higher when milestones are attempted independently from canonical snapshots, whereas the best reported continuous-evolution score is 38.03\%~\citep{deng2026swemilestone}. On scikit-learn, one reported agent--model configuration drops from 93.2\% independently to 21.1\% continuously. This is not yet the compositional prediction in Equation~\ref{eq:expected}: independent milestone scores must still be composed along the relevant task graph, with compatibility assumptions made explicit. It nevertheless shows that reset-state and persistent-rollout protocols estimate materially different quantities.

\subsection{The Horizon Residual}

Let $\Pobserved$ denote success under the natural end-to-end rollout. For a fixed declared counterfactual protocol, we summarize the comparison as

\begin{equation}
  \GammaH = \log\frac{\Pexpected}{\Pobserved},
  \label{eq:gap}
\end{equation}

where $\log$ denotes the natural logarithm, so $\GammaH$ is measured in nats. We refer to $\GammaH$ as the \emph{horizon residual}: a protocol-specific log contrast, not a direct measure of a horizon mechanism. In finite samples, it should be reported with raw counts, uncertainty, and an explicit treatment of zero or near-zero estimated probabilities. Relative to the specified counterfactual:

\begin{itemize}
  \item $\GammaH>0$ indicates lower success in the natural rollout than the model predicts;
  \item $\GammaH\approx0$ indicates agreement at the resolution supported by the data;
  \item $\GammaH<0$ indicates higher success in the natural rollout than the model predicts.
\end{itemize}

Section~\ref{sec:interpreting} gives the joint reading of these cases with both rates' magnitudes.

When the natural rollout can be evaluated at the same semantic boundaries, the contrast admits a useful stage-wise decomposition. Let $r_i=\Pr(S_i=1\mid S_{<i}=1,N)$ denote conditional stage success under the natural-rollout protocol $N$. If success requires every stage on a fixed path, then

\begin{equation}
  \Pobserved=\prod_i r_i,
  \qquad
  \GammaH=\sum_i \log\frac{q_i}{r_i}.
  \label{eq:stagewise-gap}
\end{equation}

This breakdown shows which step in the natural rollout had the largest drop relative to the checkpoint baseline. A positive term marks a stage at which success is lower in the natural rollout. It does not identify why the difference occurs. Moreover, a positive term at stage $i$ marks where the mismatch \emph{surfaces}, not necessarily where it \emph{originates}: degradation accumulated earlier may become visible only at stage $i$. Estimating later $r_i$ can also be expensive because only rollouts with successful prefixes contribute to the corresponding probability. These nested samples are correlated across stages, so intervals for the stage-wise terms should be obtained by bootstrap over complete task units rather than from the analytic approximation below (Equation~\ref{eq:gap-variance}). Branching paths and recoverable failures again require a transition model rather than this fixed-path factorization.

The four-repair example of Section~\ref{sec:motivation} makes the computation concrete. Matched checkpoints that give $q_i=0.8$ on each repair yield $\Pexpected=0.8^4\approx0.41$; a natural rollout that succeeds $10\%$ of the time then gives $\GammaH=\log(0.41/0.10)\approx1.41$~nats. The natural rollout underperforms the composed local prediction by a factor of about four. Stage-wise terms of, say, $(0.2,0.5,0.6,0.1)$ would indicate that the mismatch surfaces mainly at the third repair and would target the first intervention there, subject to the surfacing-versus-origin caveat above.

A first-order uncertainty calculation further clarifies the data requirement. Suppose $\widehat q_i$ is estimated from $m_i$ independent Bernoulli trials and $\widehat\Pobserved$ from $m_0$ independent end-to-end trials. Ignoring covariance, the delta method gives

\begin{equation}
  \operatorname{Var}(\widehat\GammaH)
  \approx
  \sum_i \frac{1-q_i}{m_i q_i}
  + \frac{1-\Pobserved}{m_0\Pobserved}.
  \label{eq:gap-variance}
\end{equation}

Thus, uncertainty grows rapidly when any local or end-to-end success probability approaches zero. In realistic evaluations, estimates may share tasks, seeds, checkpoints, or agent states, violating the independence approximation in Equation~\ref{eq:gap-variance}. Such sharing typically induces positive covariance between checkpoint and rollout estimates on the same task, so the independence approximation understates the variance of $\widehat\GammaH$. We therefore recommend paired or hierarchical bootstrap intervals over complete task units, with the analytic expression used only as intuition and a planning approximation.

The residual is diagnostic, but it is not by itself a causal explanation. A positive value marks a mismatch with the selected local-performance model, not the reason for that mismatch. If history-induced degradation is the hypothesis, the benchmark can rerun the same task with periodic context reset or compression while holding other protocol features fixed. Each intervention should define its own $\Pexpected(I)$ and $\Pobserved(I)$ before residuals are compared; keeping the original counterfactual fixed instead answers a different deployment question. A selective change in the contrast is more focused evidence than another natural-rollout curve, although remaining intervention confounds still require analysis.

\subsection{Interpreting the Residual}
\label{sec:interpreting}

The expected and observed success rates answer distinct questions and should be interpreted jointly, always relative to the specified counterfactual:

\begin{itemize}
  \item \textbf{Both are high:} both protocols place the task within the system's measured capability under the stated budgets and information.
  \item \textbf{Both are low:} the task may remain a useful deployment stress test, but the comparison has little power to separate counterfactual-model error, compounding, and history-dependent effects.
  \item \textbf{Expected success is high but observed success is low:} the natural rollout has an unexplained deficit under the selected model, motivating targeted interventions and sensitivity checks rather than an immediate mechanism claim.
  \item \textbf{Observed success exceeds expected success:} the natural rollout may exploit feedback, redundancy, adaptation, or recovery omitted by the counterfactual; interventions are still needed to determine which explanation, if any, accounts for the advantage.
\end{itemize}

If Equation~\ref{eq:expected} predicts near-zero success, the log contrast is statistically unstable and an all-failure outcome is diagnostically weak. Informative benchmark regimes should keep the counterfactual prediction away from degenerate extremes where feasible, report partial progress and cost, and include interventions that test why the two protocols differ.

\subsection{Choosing the Baseline}

The appropriate baseline depends on the scientific question. Fresh context tests a different estimand from history-matched checkpoints, which may be more relevant to deployment forecasting because they retain the expected burden of long context. State repair, as in ChainSWE's oracle mode, keeps the position in the task sequence while replacing earlier agent modifications with verified ones~\citep{jin2026chainswe}; oracle reset may additionally replace other aspects of state. Revealing a stage goal changes the information available to the agent and should not be treated as a pure state reset. No single baseline serves every estimand. Benchmarks should therefore declare checkpoint state, state-repair policy, visible history, revealed information, and local budget because each choice changes the meaning of $\Pexpected$ and $\GammaH$.

\section{Separating Task Structure from Local Difficulty}

\textbf{The key contrast is concrete:} one hundred independent renames involve much work but little dependency, while a schema--storage--API--client migration may have fewer stages but require each stage to build on the last. The number of turns does not capture this difference: one shell command may modify hundreds of files, whereas a difficult localization problem may consume many turns without adding semantic work.

We therefore advocate reporting two complementary annotations:

\begin{itemize}
  \item \textbf{Work exposure $\Nwork$:} how many required, separately verifiable stages the benchmark declares. More stages create more opportunities for local failure.
  \item \textbf{Dependency depth $\Hdep$:} how many stages lie on the deepest required chain. Greater depth means that later work must build on more earlier state or feedback.
\end{itemize}

Neither $\Nwork$ nor $\Hdep$ is an intrinsic property of a task: both depend on the chosen decomposition, acceptance conditions, and dependency graph. They can nevertheless vary independently within a declared design. Repetitive work can also be structurally shallow while becoming dependent during execution through a shared context window.

HORIZON makes a related distinction by defining an agent-independent intrinsic horizon $H^*$ as the minimum number of effective actions required by an optimal policy and by constructing longer task families through depth and breadth extensions ~\citep{wang2026mirage}. This is a useful safeguard against calling a short task ``long-horizon'' merely because an agent repeats failed actions. At the same time, $H^*$ requires an oracle, expert demonstration, or formal task specification, and its operational mapping differs across domains. We therefore view intrinsic horizon, annotated work exposure, and dependency depth as complementary descriptions: the first estimates minimal action length, while the latter two expose how a declared evaluation decomposes and connects verifiable work.

Local difficulty should likewise be measured rather than declared. A task is not simply ``hard''; it is hard \emph{for a particular agent operating with particular information, tools, verifiers, and budgets}. The same refactor may be easy when a failing test localizes the problem and difficult when feedback is delayed until final integration. For this reason, measured local success is more informative than a human-assigned easy/hard label.

Human completion time is useful for deployment calibration but does not resolve this confound. \citet{kwa2025longtasks}, for example, define the task-completion time horizon at a 50\% success rate and estimate a horizon of roughly 50 minutes for a frontier system on their task mixture. The metric communicates the scale of work that a system can reliably complete, but two tasks with the same human duration may differ substantially in annotated work exposure, dependency depth, verifier quality, and agent-local difficulty. Time horizon and the protocol-specific structural annotations proposed here therefore serve complementary purposes. Table~\ref{tab:design} sketches the factorial conditions that these two annotations induce.

\begin{table}[H]
\caption{A factorial coding-task design that separates dependency depth from measured local-stage difficulty. Work exposure and local success should be matched within each intended comparison.}
\label{tab:design}
\centering
\small
\setlength{\tabcolsep}{4pt}
\begin{tabularx}{\linewidth}{>{\raggedright\arraybackslash}p{0.17\linewidth}XX}
\toprule
 & \textbf{Easier local stages} & \textbf{Harder local stages} \\
\midrule
\textbf{Shallow chain} &
Independent repairs with diagnostic tests &
Independent repairs with ambiguous failures or security constraints \\
\textbf{Deep chain} &
An explicit migration with strong intermediate checks &
A multi-module migration with sparse feedback and difficult recovery \\
\bottomrule
\end{tabularx}
\end{table}

Task length and local difficulty will often be correlated in naturally collected tasks. They need not be statistically independent, but a benchmark can attempt to separate them experimentally by varying the declared dependency structure while approximately matching local success, and by varying local difficulty while holding that structure fixed.

\section{A Benchmark Design Protocol for Coding and Terminal Agents}
\label{sec:protocol}

Coding and terminal environments are particularly well suited to this design. Repository snapshots and containers make earlier states replayable, while tests and state invariants provide semantic acceptance conditions. We propose the following compact workflow:

\begin{enumerate}
  \item \textbf{Pre-register.} Declare the task, stages, acceptance conditions, dependency graph, rollout policy, tools, verifier, and recovery rules before evaluation.
  \item \textbf{Define checkpoints.} Record the environment state, visible history, revealed information, repair policy, and rules for alternative valid implementations.
  \item \textbf{Match budgets.} Report token, tool, time, verification, and retry costs for both local and end-to-end runs. Where possible, include both equal per-stage budgets and an aggregate-budget-matched comparison; otherwise count the resource difference as part of the intervention.
  \item \textbf{Audit local performance.} Report per-stage successes, trial counts, uncertainty, and heterogeneity; pre-specify how zero or near-zero estimates will be handled.
  \item \textbf{Choose the model.} Use the product baseline for a fixed compatible path. Use a conditional model or checkpoint simulator when the task includes branches, retries, recovery, or shared-state dependence.
  \item \textbf{Run and intervene.} Estimate $\Pexpected$ before measuring $\Pobserved$, then test targeted changes such as history reset, state repair, stronger verification, rollback, or explicit planning.
  \item \textbf{Test sensitivity.} Recompute predictions and residuals under multiple reasonable decompositions, dependency graphs, and admissible checkpoint protocols. Call a conclusion robust only when its sign is consistent across the pre-declared set; otherwise report the range of $\GammaH$.
  \item \textbf{Audit the benchmark.} Report task-inclusion criteria, exclusions, leakage checks, and contamination risks for both the agent and any decomposition or grading procedure.
\end{enumerate}

An informative task family should contain both shallow and deep variants. Independent edits provide a high-exposure, shallow control. A causally ordered patch stack (for example, schema, then storage, then API, then client) provides a deep condition. The local stages should be selected so that their measured success distributions are comparable. Both stage-given and goal-only variants are useful because a checkpoint may reveal localization or decomposition information that is unavailable in the natural task.

History-induced degradation also warrants a dedicated controlled comparison, matching the useful control of Table~\ref{tab:three-reasons}. The agent can perform the same sequence of simple operations while retaining the full history, periodically compressing it, or restarting from a clean context. Such a comparison helps estimate the total effect of the declared history intervention; it does not identify a single underlying mechanism. Observational evidence underscores why the controlled version matters: in an ultra-long-horizon software benchmark, none of the 71 rollouts that used mid-trajectory context compaction passed, against 8.9\% without it~\citep{desai2026swemarathon}; absent a matched intervention, such a correlation cannot distinguish compaction as a cause of failure from compaction as a marker of rollouts that were already failing. Environment state, revealed information, and budgets must be held fixed across conditions unless their joint intervention is the stated estimand.

\section{Related Evidence}
\label{sec:related}

\begin{figure}[H]
  \centering
  \makebox[\linewidth][c]{\includegraphics[width=0.98\linewidth]{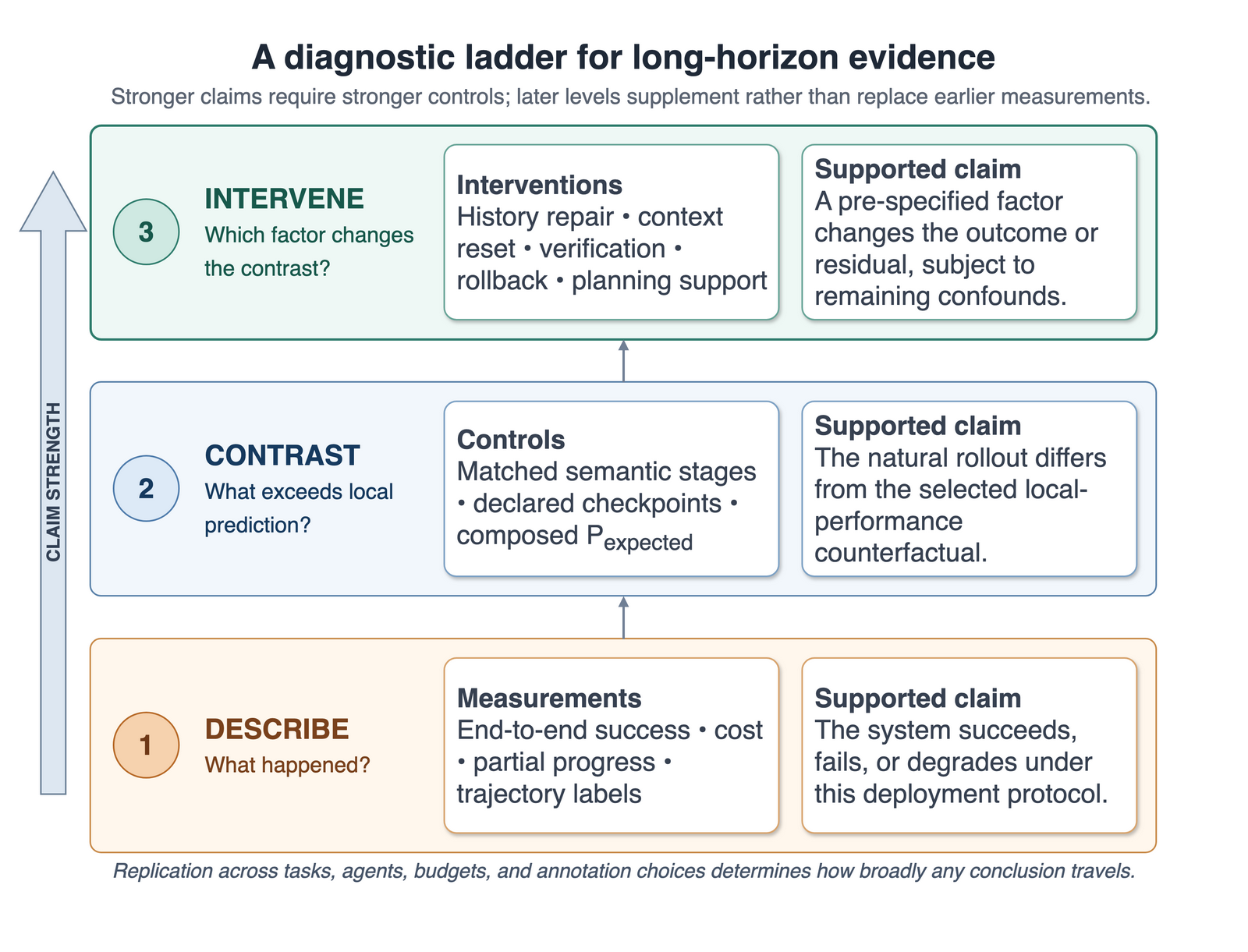}}
  \caption{Hierarchy of evidence for long-horizon evaluation. Descriptive measurements characterize deployment behavior; a declared local-stage counterfactual identifies mismatch with matched competence; and targeted interventions test scoped mechanism hypotheses. Replication across tasks, systems, budgets, and annotation choices determines the generality of each conclusion.}
  \label{fig:diagnostic-ladder}
\end{figure}

The literature suggests a hierarchy of evidential strength. End-to-end curves answer what happened under a deployment protocol; trajectory labels and dense rewards add localization~\citep{lightman2024lets}; matched checkpoints test whether the natural rollout exceeds local prediction; and targeted interventions test whether a proposed factor changes that contrast~\citep{sinha2026illusion,wang2026mirage}. Figure~\ref{fig:diagnostic-ladder} summarizes this progression. The levels are cumulative: mechanism experiments do not remove the need to report raw success, cost, and partial progress.

\begin{table}[tbp]
\caption{Representative empirical evidence, the history or state control implemented by each study, and the diagnostic question that remains unresolved. Reported values follow study-specific protocols and are not directly comparable across rows.}
\label{tab:evidence}
\centering
\scriptsize
\setlength{\tabcolsep}{3pt}
\begin{tabularx}{\linewidth}{>{\raggedright\arraybackslash}p{0.16\linewidth}
>{\raggedright\arraybackslash}p{0.29\linewidth}
>{\raggedright\arraybackslash}p{0.20\linewidth}>{\raggedright\arraybackslash}X}
\toprule
\textbf{Study} & \textbf{Reported evidence} & \textbf{History/state control} &
\textbf{What remains unresolved?} \\
\midrule
SWE-EVO~\citep{le2025sweevo} &
48 release-sized tasks modify 20.9 files on average and are checked by 874 tests on average. &
Each task starts from its prescribed release state. &
Its reported protocol is not designed to separate scale, local difficulty, and horizon. \\
NL2Repo-Bench~\citep{ding2025nl2repo} &
Across 104 repository-generation tasks, the best reported mean test pass rate is 40.2\%; no more than five repositories are fully completed. &
One persistent workspace per task. &
The reported protocol does not include a matched-stage prediction to distinguish error compounding from failures of cross-stage coherence. \\
SWE-Chain~\citep{lam2026swechain} &
12 release chains contain 155 transitions and 1,660 grounded requirements. &
Code persists across transitions; conversations reset. &
The reported transition estimates are not composed into a chain-level counterfactual. \\
ChainSWE~\citep{jin2026chainswe} &
Per-bug accuracy is 58.9\% with oracle-correct prior state and 36.5\% with agent-generated prior state. &
Direct oracle-versus-persistent-state comparison. &
The reported comparison does not compose local estimates into an expected chain outcome or test multiple mechanism-specific interventions. \\
SlopCodeBench~\citep{orlanski2026slopcodebench} &
No evaluated agent completes any of the 36 problems; erosion and verbosity increase in 77\% and 75.5\% of trajectories, respectively. &
Agents repeatedly extend their own solutions. &
The reported protocol characterizes quality degradation without a matched reset-state baseline. \\
SWE-Marathon~\citep{desai2026swemarathon} &
Across 20 project-scale tasks and 1,300 rollouts, no configuration exceeds 30\% pass@1; 0 of 71 compaction-path trials pass versus 8.9\% without. &
Persistent multi-hour rollouts with post-hoc trajectory audits. &
Compaction and repetition correlate with failure but are not intervened on, so cause and marker remain confounded. \\
SWE-Milestone~\citep{deng2026swemilestone} &
Independent milestone scores exceed roughly 80\%; the best continuous score is 38.03\%. &
Canonical snapshots versus a persistent codebase. &
The reported independent scores are not composed along semantic prerequisite paths. \\
Illusion~\citep{sinha2026illusion} &
At a matched evaluation turn, increasing the error rate in a fixed-format prior history reduces subsequent accuracy. &
Artificially healed or corrupted histories. &
The controlled synthetic task does not establish prevalence or effect size in open-ended agents. \\
HORIZON~\citep{wang2026mirage} &
More than 3,100 trajectories across four domains reveal distinct planning-, memory-, and history-related failure patterns. &
Controlled depth or breadth extensions within task families. &
Post-hoc, potentially co-occurring labels describe failure composition rather than causal mechanisms. \\
\bottomrule
\end{tabularx}
\end{table}

Table~\ref{tab:evidence} highlights how prior benchmarks observe performance drops on long tasks but do not yet combine all the history and state controls needed to isolate why the drop occurred. The studies differ in whether they control local solvability, repair prior state, or intervene on history.

\begin{table}[tbp]
\caption{Component-level delta between representative studies and the proposed protocol, judged from each study's reported protocol. \checkmark: implemented; $\sim$: partial or implicit; ---: not part of the reported protocol.}
\label{tab:delta}
\centering
\scriptsize
\setlength{\tabcolsep}{3pt}
\begin{tabularx}{\linewidth}{>{\raggedright\arraybackslash}p{0.17\linewidth}*{6}{>{\centering\arraybackslash}X}}
\toprule
\textbf{Study} & \textbf{Matched local stages} & \textbf{Composed $\Pexpected$} & \textbf{Natural rollout} & \textbf{State compatibility} & \textbf{Targeted intervention} & \textbf{Protocol sensitivity} \\
\midrule
SWE-EVO & --- & --- & \checkmark & --- & --- & --- \\
NL2Repo-Bench & --- & --- & \checkmark & --- & --- & --- \\
SWE-Chain & $\sim$ & --- & \checkmark & --- & --- & --- \\
ChainSWE & \checkmark & --- & \checkmark & $\sim$ & $\sim$ & --- \\
SlopCodeBench & --- & --- & \checkmark & --- & --- & --- \\
SWE-Marathon & --- & --- & \checkmark & --- & --- & --- \\
SWE-Milestone & \checkmark & --- & \checkmark & $\sim$ & --- & --- \\
Illusion & \checkmark & $\sim$ & $\sim$ & --- & \checkmark & --- \\
HORIZON & --- & --- & \checkmark & --- & $\sim$ & --- \\
\midrule
This position (proposed) & \checkmark & \checkmark & \checkmark & \checkmark & \checkmark & \checkmark \\
\bottomrule
\end{tabularx}
\end{table}

Table~\ref{tab:delta} restates this comparison as a component-level audit of the proposed sequence. Every component exists in some published evaluation, and we claim no novelty for any of them individually. The position is that no reviewed study implements the full sequence (matched local measurement, declared composition, natural rollout, state-compatibility handling, targeted intervention, and protocol-sensitivity analysis) under one pre-specified protocol. Partial marks record, for example, ChainSWE's single oracle-repair intervention, SWE-Milestone's inferred rather than declared dependency graph, and the homogeneous $p^n$ analysis of \citet{sinha2026illusion} on a controlled synthetic task rather than a natural deployment rollout. If a prior evaluation is found to implement the complete sequence, the claim of this paper narrows to advocacy of its standardization.

\paragraph{From isolated issue resolution to sustained software evolution.}
SWE-bench established repository-level issue resolution as a practical testbed for language models~\citep{jimenez2024swebench}; SWE-Gym and OpenHands contributed executable training tasks, trajectories, and a general agent platform~\citep{pan2025swegym,wang2025openhands}. SWE-Lancer and time-calibrated evaluations broaden ecological validity through freelance work and human-duration estimates~\citep{miserendino2025swelancer,kwa2025longtasks}. SWE-Bench Pro, SWE-EVO, and NL2Repo-Bench extend scope toward enterprise issues, release-sized changes, and full-repository construction~\citep{deng2025swebenchpro, le2025sweevo,ding2025nl2repo}. Their results indicate that current systems struggle as scope and coordination demands grow. Their reported protocols are not designed to hold local difficulty, planning burden, feedback sparsity, and work exposure fixed while varying horizon, so they motivate rather than instantiate the counterfactual proposed here.

\paragraph{Agent scaffolding as part of the evaluated system.}
The deployed agent is more than its base model. SWE-agent shows that an agent-computer interface can alter repository navigation, editing, and testing behavior ~\citep{yang2024sweagent}. AutoCodeRover instead uses program structure and fault localization to focus search~\citep{zhang2024autocoderover}; Agentless replaces open-ended tool planning with localization, repair, and patch validation~\citep{xia2024agentless}; and MASAI delegates reproduction, localization, repair, and ranking to specialized modules ~\citep{arora2024masai}. Their reported results use different models, prompts, costs, and benchmark versions, so they should not be read as a direct leaderboard. Collectively, they show why local-stage and natural-rollout evaluations must preserve the same agent configuration: changing the scaffold changes the capability being measured.

\paragraph{Stateful interactive environments beyond software repositories.}
The same concern appears outside software maintenance. WebArena evaluates realistic web workflows and reports 14.41\% success for its best GPT-4-based agent versus 78.24\% for humans~\citep{zhou2023webarena}. OSWorld evaluates 369 open-ended computer tasks and reports 12.24\% for its best evaluated model versus 72.36\% for humans ~\citep{xie2024osworld}. AppWorld combines nine applications and 457 APIs with state-based tests that also detect collateral changes~\citep{trivedi2024appworld}. RE-Bench complements binary completion with continuously scored research-engineering objectives and varying time budgets; agents lead at short total budgets, while human experts show stronger returns to additional time~\citep{wijk2024rebench}. These benchmarks improve realism and stateful evaluation, but low success alone still mixes grounding, operational knowledge, planning, local difficulty, and horizon. Table~\ref{tab:broader-evidence} summarizes these system-level factors alongside the scaffolding studies above.

\begin{table}[tbp]
\caption{Evidence that interface design, agent scaffolding, environment state, and resource allocation are constitutive components of the evaluated system. Results follow study-specific protocols and are not directly comparable across rows.}
\label{tab:broader-evidence}
\centering
\scriptsize
\setlength{\tabcolsep}{3pt}
\begin{tabularx}{\linewidth}{>{\raggedright\arraybackslash}p{0.19\linewidth}
>{\raggedright\arraybackslash}p{0.31\linewidth}X}
\toprule
\textbf{Study} & \textbf{Design or reported signal} & \textbf{Relevance to this position} \\
\midrule
SWE-agent~\citep{yang2024sweagent} &
Custom agent-computer interface; 12.5\% pass@1 on full SWE-bench. &
Interface design changes natural interaction behavior and must be matched across conditions. \\
AutoCodeRover~\citep{zhang2024autocoderover} &
AST-guided search and optional fault localization; 19\% on SWE-bench Lite. &
Search and localization support alter the effective local difficulty. \\
Agentless~\citep{xia2024agentless} &
Fixed localization--repair--validation pipeline; 32.0\% on SWE-bench Lite. &
A controlled composition of stages can compete with open-ended trajectories. \\
MASAI~\citep{arora2024masai} &
Specialized sub-agents; 28.33\% on SWE-bench Lite. &
Modular decomposition changes context exposure and coordination burden. \\
WebArena~\citep{zhou2023webarena} &
Realistic web workflows; 14.41\% best-agent versus 78.24\% human success. &
End-to-end failure combines horizon with grounding, planning, and website state. \\
OSWorld~\citep{xie2024osworld} &
369 desktop and web tasks; 12.24\% best-model versus 72.36\% human success. &
Persistent operating-system state broadens the sources of trajectory failure. \\
AppWorld~\citep{trivedi2024appworld} &
750 tasks across nine apps and 457 APIs, graded by state-based tests. &
State checks can expose partial correctness and unintended collateral changes. \\
RE-Bench~\citep{wijk2024rebench} &
Seven research-engineering environments with continuous scores and multiple budgets. &
Capability conclusions depend on time allocation and best-of-$k$ versus single-trajectory protocols. \\
\bottomrule
\end{tabularx}
\end{table}

\paragraph{Partial-progress measurement and persistent-state controls.}
Terminal success can hide where a rollout failed. LHTB uses deterministic subtask grading, so runs with meaningful partial reward are distinguishable from both passing runs and no-progress failures~\citep{li2026lhtb}. HORIZON analyzes more than 3,100 trajectories across web, operating-system, database, and embodied domains with a seven-category failure taxonomy~\citep{wang2026mirage}. Its trajectory-grounded judge reaches $\kappa=0.84$ against one human annotator on a 40-trajectory pilot, supporting scalable descriptive annotation. The labels may co-occur and are assigned after observing the rollout, however, so agreement supports reproducibility of attribution rather than causal identification. SWE-Chain, ChainSWE, and SlopCodeBench instead preserve evolving software state across transitions or checkpoints~\citep{lam2026swechain,jin2026chainswe, orlanski2026slopcodebench}. Among the studies reviewed here, ChainSWE additionally reports a direct oracle-state comparison. These designs expose error propagation and architectural degradation that isolated issue protocols reset away, while leaving open how matched local performance should be composed into an expected end-to-end outcome.

\paragraph{Checkpoint composition and mechanism-targeted interventions.}
SWE-Milestone closely aligns with our proposed direction by comparing canonical-snapshot milestone execution with continuous execution on a persistent codebase ~\citep{deng2026swemilestone}. Its inferred graph is useful for evaluating history and patch execution, but it need not be a minimal graph of semantic prerequisites: an alternative valid implementation may not reproduce the reference path. This illustrates why dependency annotations and state compatibility must be explicit when composing local estimates. More direct mechanism evidence comes from intervention. By manipulating errors in a fixed-format prior history, \citet{sinha2026illusion} estimate sensitivity to prior errors at a matched evaluation turn, while noting that the synthetic result is not sufficient for real-world long-horizon execution. Broader task ontologies and controlled horizon variation provide complementary ingredients~\citep{dong2026survey,kim2026training}. Our position is that replayable checkpoints, dense progress measures, persistent-state comparisons, and targeted interventions should be combined under a pre-specified, agent-configuration-matched counterfactual, a combination that Table~\ref{tab:delta} shows no reviewed study yet implements in full.

\section{Scope, Limitations, and Falsifiability}
\label{sec:limitations}

The proposed quantities are properties of a declared evaluation protocol, not intrinsic properties of a task or model. Stage decompositions are not unique, and dependency edges can be implementation-relative. Canonical checkpoints may also reveal privileged information about decomposition, localization, or the reference solution. We therefore advocate publishing the decomposition, acceptance conditions, dependency annotations, checkpoint construction, and revealed information, and testing whether conclusions persist across multiple reasonable annotations and both stage-given and goal-only conditions.

The decomposition is also a governance problem, not only a technical one. Someone must author it, and its quality bounds every downstream quantity. We recommend that decompositions be authored or reviewed by domain experts, versioned alongside the benchmark, and arbitrated through the annotation-sensitivity tests of Section~\ref{sec:protocol} rather than by fiat; LLM-assisted decomposition is acceptable when each stage's acceptance condition is executed against ground truth, since an unverifiable stage silently redefines the estimand.

Statistical and resource assumptions are equally consequential. Local outcomes may be heterogeneous or correlated across tasks, stages, agents, and random seeds; separately started stages may receive more total search, verification, or retry budget than a natural rollout. Benchmark reports should therefore include raw counts, uncertainty, resource and cost accounting, conditional or simulation-based alternatives to the product model, and predeclared rules for zero and near-zero probabilities. Verifier error belongs in the same ledger: an agentic audit of graded rollouts reports judge--verifier disagreement of 1.4\% for hand-written functional verifiers versus 32.4\% for inherited pull-request tests~\citep{huang2026deepswe}; at the latter rate, measurement noise alone can dominate a small residual, so residual estimates should report verifier validation alongside raw counts. Developing estimators that propagate uncertainty through branching state graphs is an important part of the empirical agenda.

A further risk is protocol sensitivity: two admissible checkpoint protocols can in principle yield opposite-signed residuals on the same task family. The protocol treats this scenario as a result rather than a failure. It indicates that no robust conclusion exists at the chosen resolution, but it also motivates the robustness criterion of Section~\ref{sec:protocol}: conclusions should be reported only when their sign is stable across the pre-declared admissible set, and as protocol ranges otherwise.

Finally, replayable and separately verifiable stages favor coding and terminal tasks with strong infrastructure. Task inclusion, public-repository exposure, reference-derived checkpoints, and verifier availability can introduce selection or contamination bias. The framework's value outside these domains remains an empirical question. Most importantly, $\GammaH$ does not identify context degradation, planning failure, state contamination, or recovery; it identifies a contrast under one protocol.

\paragraph{Falsifiers of this position.}
Our position would be substantially weakened if any of the following were demonstrated:
\begin{enumerate}
  \item across multiple pre-declared, admissible protocols on the same task family, the sign of $\GammaH$ is unstable in ways that admissibility criteria cannot resolve;
  \item compatible checkpoints cannot be constructed at manageable cost for representative coding benchmarks; or
  \item the residual provides no incremental diagnostic information beyond existing persistent-versus-reset comparisons across several benchmark families.
\end{enumerate}
We invite empirical work targeting these tests.

\section{Anticipated Objections and Research Agenda}
\label{sec:agenda}

\paragraph{Q: Isn't end-to-end success what deployment cares about?}
\textbf{A:} Yes. Deployment and diagnostic benchmarks answer different questions, and neither replaces the other. Our claim applies only when a deployment result is used to argue for a particular failure mechanism.

\paragraph{Q: Won't scaling dissolve the problem?}
\textbf{A:} Longer contexts and better base models may reduce the residual, but they do not directly address state contamination, planning, or recovery. If a scaled system reaches $\GammaH\approx0$ under a declared baseline, that is exactly the evidence this position asks for.

\paragraph{Q: Don't standard ablations already identify mechanisms?}
\textbf{A:} Ablations measure component contributions within a fixed protocol. The residual asks whether matched local performance predicts the end-to-end result. The two are complementary: the residual locates a mismatch that an ablation can then help explain.

\paragraph{Q: If everything depends on the decomposition, how can results be compared?}
\textbf{A:} Publish, replicate, and vary the decomposition. The robustness criterion in Section~\ref{sec:protocol} withholds conclusions that do not survive reasonable alternatives. Declared dependence can be audited; hidden dependence cannot.

The objections above mark the boundaries of the claim; the following agenda moves within them:
\begin{enumerate}
  \item \textbf{Instantiate the comparison.} Apply the protocol to one existing persistent-state benchmark and report $\Pexpected$, $\Pobserved$, and $\GammaH$ with bootstrap intervals, reusing published decompositions where available.
  \item \textbf{Develop branching estimators.} Extend the uncertainty treatment of Equation~\ref{eq:gap-variance} to conditional transition models over branching state graphs, with explicit rules for zero and near-zero probabilities.
  \item \textbf{Run the intervention program.} Pair each recurring residual pattern with single-factor interventions, history reset or compression, state repair, verification, rollback, explicit planning, so that residual patterns become testable causal hypotheses rather than labels.
  \item \textbf{Map protocol robustness.} Characterize when admissible protocols agree in sign and when they diverge, turning the robustness criterion of Section~\ref{sec:protocol} into empirically grounded admissibility rules.
  \item \textbf{Transfer beyond coding.} Identify where replayable state and semantic acceptance conditions exist outside coding and terminal tasks, and adapt the protocol where they do not.
\end{enumerate}

We close with the strongest form of the claim. A declining end-to-end curve licenses a claim about deployment; a claim about long-horizon \emph{mechanisms} should be licensed only by an evaluation that reports, alongside end-to-end success, a pre-specified compositional prediction and the resulting residual.

\section{Conclusion}

Long tasks are globally harder in part because they require more work. Our position is that the central scientific question is whether natural end-to-end rollouts perform differently from a pre-specified, intervention-specific model of local competence. We therefore advocate pre-specifying the stage decomposition and intervention protocol, estimating a protocol-specific counterfactual, measuring the natural rollout, and using targeted interventions to investigate the resulting contrast. The horizon residual is not a causal answer; it is a disciplined starting point for testing which aspects of history, state, information, and recovery make long-horizon execution differ. Until such residuals are reported, a declining curve should be read as evidence about deployment, not about mechanism. This framing turns ``longer is harder'' from a descriptive observation into an auditable research program.

\clearpage
\bibliographystyle{plainnat}
\bibliography{references}

\end{document}